\documentclass{article}

\PassOptionsToPackage{numbers, compress}{natbib}

\usepackage[final]{neurips_wrl2022}

\usepackage[utf8]{inputenc} 
\usepackage[T1]{fontenc}    
\usepackage{hyperref}       
\usepackage{url}            
\usepackage{booktabs}       
\usepackage{amsfonts}       
\usepackage{nicefrac}       
\usepackage{microtype}      
\usepackage{array,multirow,graphicx}
\usepackage[linesnumbered,ruled,vlined]{algorithm2e}
\usepackage[font=small,labelfont=bf]{caption}
\usepackage{amsmath}
\usepackage{amssymb}
\usepackage[table]{xcolor}
\usepackage{paralist} 
\usepackage{subcaption}
\usepackage[nameinlink]{cleveref}

\newcommand{\MONTH}{%
	\ifcase\month
	\or January
	\or February
	\or March
	\or April
	\or May
	\or June
	\or July
	\or August
	\or September
	\or October
	\or November
	\or December
	\fi}

\newcommand{\Dataset}{\mathcal{D}}

\newcommand{\cliport}{\textsc{CLIPort}}

\DeclareMathOperator*{\argmax}{arg\,max}

\newcommand{\ctab}[1]{\begin{tabular}{@{}c@{}}#1\end{tabular}}

\title{PARTNR: Pick and place Ambiguity Resolving by Trustworthy iNteractive leaRning}

\author{%
    Jelle Luijkx $\quad$ Zlatan Ajanović $\quad$ Laura Ferranti $\quad$ Jens Kober\\
    Department of Cognitive Robotics\\
    Delft University of Technology, The Netherlands\\
    \texttt{\{J.D.Luijkx, Z.Ajanovic, L.Ferranti, J.Kober\}@tudelft.nl} \\[1em]
    \large\textbf{\href{https://partnr-learn.github.io/}{\url{partnr-learn.github.io}}}
}

\begin{document}

\maketitle

\begin{abstract}

Several recent works show impressive results in mapping language-based human commands and image scene observations to direct robot executable policies (e.g., pick and place poses).
However, these approaches do not consider the uncertainty of the trained policy and simply always execute actions suggested by the current policy as the most probable ones. This makes them vulnerable to domain shift 
and inefficient in the number of required demonstrations. 
We extend previous works and present the PARTNR algorithm that can detect ambiguities in the trained policy by analyzing multiple modalities in the pick and place poses using topological analysis. PARTNR employs an adaptive, sensitivity-based, gating function that decides if additional user demonstrations are required.
User demonstrations are aggregated to the dataset and used for subsequent training.
In this way, the policy can adapt promptly to domain shift and it can minimize the number of required demonstrations for a well-trained policy. The adaptive threshold enables to achieve the user-acceptable level of ambiguity to execute the policy autonomously and in turn, increase the trustworthiness of our system.
We demonstrate the performance of PARTNR in a table-top pick and place task.

\end{abstract}

\section{Introduction}


Despite the numerous exciting results in robot learning, only a few methods are actually robust enough to be employed in everyday life.
Many manipulation tasks, such as pick-and-place in household scenarios, are challenging for robots, while they are actually easy for humans.
To overcome this performance mismatch, we can exploit the human domain knowledge through (interactive) imitation learning \cite{celemin2022interactive}.
This requires novel methods with an intuitive interface to transfer non-expert user knowledge to robotic systems. 
The impressive capabilities of recently introduced foundation models can possibly ease this transfer of knowledge.
Foundation models can be trained on language data only (e.g., Transformers \cite{vaswani2017attention}, BERT \cite{devlin2018bert}, or GPT-3 \cite{brown2020language}) or can be trained on multi-modal data, such as images and their captions (e.g., CLIP \cite{radford2021learning}).
In the field of robotics, language foundation models can be used for task planning \cite{ahn2022can} and interpreting human commands \cite{shridhar2021cliport, ahn2022visually} as well as corrections \cite{sharma2022correcting}.
In particular, in the setting of (interactive) imitation learning, it is a natural choice to exploit language foundation models, since it allows the user to give instructions or corrections in an intuitive manner.
Interactive imitation learning is a subclass of imitation learning in which the human is influencing the learning loop while executing the task \cite{celemin2022interactive}.
To be practical and trustworthy, the robot should ask for help when it is uncertain about the outcome of its actions.
At the same time, humans should not be bothered too much. 
In this work, we address this problem by introducing Pick and place Ambiguity Resolving by Trustworthy iNteractive leaRning (PARTNR).
Our work is related to the seminal work that introduced dataset aggregation (DAgger) for imitation learning \cite{ross2011reduction}.
DAgger addresses the compounding errors in imitation learning caused by covariate shift through the collection of on-policy data.
Many variants were introduced afterward, such as Human-Gated DAgger (HG-DAgger) \cite{kelly2019hgdagger}, where the expert can take over control if deemed necessary, and Ensemble-DAgger\cite{menda2019ensembledagger}, where the robot queries expert input when a novel or risky situation is faced.
Regarding the ambiguity resolution, our work is related to LIRA \cite{franzese2020learning} which treats ambiguities in discrete reference frames, while here we focus on continuous actions.
PARTNR can be used to interactively train vision-based pick and place models, such as Transporter networks \cite{zeng2021transporter} and its extension for language commands CLIPort \cite{shridhar2021cliport}.
PARTNR asks for a human demonstration in case the model predictions are ambiguous. 
We consider a prediction to be ambiguous if it results in multiple options with similar value estimates.
To be trustworthy, the threshold of the gating function is adaptive and allows it to satisfy a user-defined sensitivity, balancing between potential failing and asking the user unnecessarily.
%
PARTNR consists of two main steps:
\begin{inparaenum}[1)]
    \item Detecting ambiguity in pick and place heatmaps by finding multiple local maxima using topological persistence and query user demonstrations if needed.
    \item Aggregating data from new human demonstrations in DAgger style to learn from feedback and resolve the ambiguity.
\end{inparaenum}
PARTNR has several advantageous features compared to the other state-of-the-art approaches:
\begin{inparaenum}[1)]
    \item By querying a new demonstration based on the level of ambiguity, it avoids gathering demonstrations for situations that are already learned by the agent, therefore reducing the number of required demonstrations.
    \item By specifying the desired sensitivity level, the user can set its preferred balance between the frequency of queries by the robot and the failure rate, therefore increasing the system's trustworthiness.
    \item By gathering data during execution, PARTNR can adapt to changing environments as well as include \emph{failure states}, i.e., new states visited by making mistakes, in the dataset so it can learn to recover from them.
\end{inparaenum}
We demonstrate these on a simulated table-top robot pick and place task, where we show an improvement of the performance with respect to the baseline (CLIPort variant). 

The rest of the paper is organized as follows. Section \ref{sec:prel} presents preliminaries as seen in the works \cite{zeng2021transporter, shridhar2021cliport} and lays the formal problem formulation for our work. Section \ref{sec:partnr} presents our method. This is followed by experiments and conclusion sections.

Additional material is available at: \href{https://partnr-learn.github.io}{partnr-learn.github.io}.

\begin{figure}[t]
    \centering
    \includegraphics[width=\linewidth]{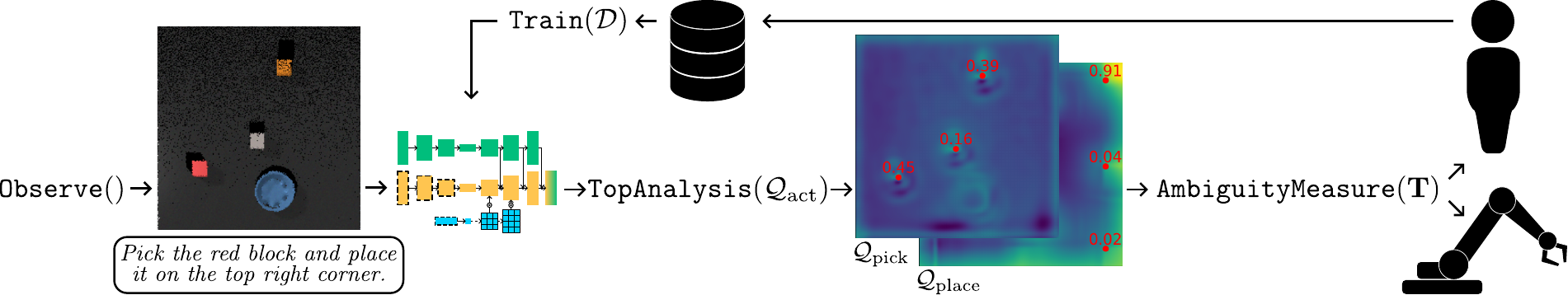}
    \captionsetup{belowskip=0pt}
    \setlength{\belowcaptionskip}{-8pt}
    \addtolength{\textfloatsep}{-0.2in}
    \caption{PARTNR framework on an example task.}
    \label{fig:framework}
\end{figure}

\section{Preliminaries and Problem Definition} \label{sec:prel}

We follow \cite{zeng2021transporter, shridhar2021cliport} and describe the pick and place problem as finding a mapping from an observation $\mathbf{o}_t$ to a pick and place action $\mathbf{a}_t$ at time step $t$, that is, 
$f(\mathbf{o}_t) \rightarrow \mathbf{a}_t = (\mathcal{T}_\mathrm{pick}, \mathcal{T}_\mathrm{place}) \in \mathcal{A}$, where $\mathcal{A}$ is the set of possible actions and $\mathcal{T}_\mathrm{pick}$ and $\mathcal{T}_\mathrm{place}$ are the end-effector pick and place poses, respectively.
We consider a table-top pick and place problem with $\mathcal{T}_\mathrm{pick}, \mathcal{T}_\mathrm{place} \in \mathbb{R}^2$.
Motion primitives can be used to obtain a sequence of lower-level actions from $\mathcal{T}_\mathrm{pick}$ and $\mathcal{T}_\mathrm{place}$.
%
Furthermore, we consider vision-based manipulation with $\mathcal{T}_i \sim (u, v), i \in \{\mathrm{pick}, \mathrm{place} \}$, where $(u, v)$ is a pixel location of a (projected) top view image.
If we model the action-value functions $\mathcal{Q}_\mathrm{pick}$ and $\mathcal{Q}_\mathrm{place}$, the optimal pick and place locations according to the model are $\mathcal{T}_\mathrm{pick} = \argmax_{(u, v)} \mathcal{Q}_\mathrm{pick}((u,v)|\mathbf{o}_t)$ and $\mathcal{T}_\mathrm{place} = \argmax_{(u, v)} \mathcal{Q}_\mathrm{place}((u,v)|\mathbf{o}_t, \mathcal{T}_\mathrm{pick})$.
Note that $\mathcal{T}_\mathrm{place}$ is conditioned on $\mathcal{T}_\mathrm{pick}$.
Normalized heatmaps correlated with pick and place success can be obtained using the $\mathrm{softmax}$ function, i.e., $\mathcal{V}_\mathrm{pick}\in\mathbb{R}^{H{\times}W}=\mathrm{softmax}(\mathcal{Q}_\mathrm{pick}((u,v)|\mathbf{o}_t))$
, where $H$ and $W$ are the height and width of the top view image, respectively. 
The action-value functions can be estimated through imitation learning.
We build on the standard imitation learning setting where we have a dataset
$\mathcal{D}=\{\zeta_1,\zeta_2, ..., \zeta_n\}$, where $n$ is the number of expert demonstration trajectories consisting of one or more tuples of observations and actions, i.e., $\zeta_i=\{(\mathbf{o}_0,\mathbf{a}_0), (\mathbf{o}_1,\mathbf{a}_1), ...\}$.

In this work, we extend previous problem formulation and consider situations when taking the most probable action by $\argmax$ is not sufficient, e.g. when there is no single distinctive maximum. We tackle the interactive learning problem where the robot needs to hand over the control back to the human and learn from new human demonstrations.

\section{PARTNR: Pick and place Ambiguity Resolving by Trustworthy iNteractive leaRning} \label{sec:partnr}

PARTNR is an interactive imitation learning algorithm that asks the human to take over control in case it considers the situation to be ambiguous. The situation is ambiguous when the learned policy does not provide a single dominant solution, i.e., there are multiple local maxima with close values in the action space. User demonstrations are aggregated to the dataset $\Dataset$ and used for subsequent training, as shown in \Cref{alg:partnr}. 
The robot observes, at each execution step, a human-provided natural language command and the state of the environment (e.g., a top-view image of the table).
Based on the observation, the policy provides the heatmap, representing the value of the action (e.g., $\mathcal{Q}_\mathrm{pick} $ representing pick location). 
The heatmap ($\mathcal{Q}_\mathrm{pick}$ and subsequently $\mathcal{Q}_\mathrm{place}$) is then analyzed to detect multiple local maxima (in \texttt{TopAnalysis}).
In this work, we rely on computational topology methods for finding local maxima \cite{edelsbrunner2010computational}. Specifically we use a persistent homology method \cite{huber2021persistent}. 
Then, in \texttt{AmbiguityMeasure}, the obtained corresponding values of the local maxima $\mathbf{T}$, are normalized using the $\mathrm{softmax}$ function and the maximum value $\hat{p}_{\mathrm{act}}$ is then used to decide if the situation is ambiguous. If $\hat{p}_{\mathrm{act}}$ is smaller than the threshold $p_{\mathrm{act}}^\mathrm{thr}$, the situation is ambiguous. In case the  situation is ambiguous, the robot is not executing the policy but queries the human teacher. The threshold $p_{\mathrm{act}}^\mathrm{thr}$ is updated continuously, at every step, by function \texttt{UpdateThreshold}, to satisfy a user-defined sensitivity value (more details in \Cref{sec:threshold}). Whenever there is a teacher input, the data is aggregated and the policy is updated using the function \texttt{Train} (like in \cite{ross2011reduction}).

\autoref{fig:framework} shows the PARTNR framework in an example where a human asks the robot to pick the red block and place it in the top right corner. As we can see on the heatmaps for pick and place, there are multiple local maxima for this command. In the pick heatmap $\mathcal{Q}_\mathrm{pick}$ there are at least three local maxima, each related to one of the blocks. The maximum related to the red block is the highest ($0.45$). However, the orange block is also relatively close ($0.39$). In this case, the situation might be ambiguous (depending on the sensitivity level) and the robot might query the teacher for a demonstration.

\section{Experiments and Results}

\begin{algorithm}[b]
\DontPrintSemicolon
\fontsize{7pt}{8pt}\selectfont
	\SetKwFunction{Observe}{Observe}
	\SetKwFunction{CLIPort}{\cliport}
	\SetKwFunction{Train}{Train}
	\SetKwFunction{QueryTeacher}{QueryTeacher}
	\SetKwFunction{TopAnalysis}{TopAnalysis}
	\SetKwFunction{AmbiguityMeasure}{AmbiguityMeasure}
	\SetKwFunction{Threshold}{UpdateThreshold}
	\SetKwFunction{ObserveCorrection}{ObserveCorrection}
	\SetKwFunction{UpdateFlags}{UpdateFlags}
	\SetKwFunction{Act}{Act}
	\SetKwInOut{Input}{input}\SetKwInOut{Output}{output}
	\Output{$\mathcal{Q}_\mathrm{pick}, \mathcal{Q}_\mathrm{place}$ \tcp*[r]{pick and place value functions} } 
	\BlankLine
	\Begin{
	
		$\Dataset, \mathcal{Q}_\mathrm{pick}, \mathcal{Q}_\mathrm{place}, p_{\mathrm{pick}}^\mathrm{thr} , p_{\mathrm{place}}^\mathrm{thr}            \gets \texttt{init}()  $\tcp*[r]{initialization}
		\BlankLine
		\For(\tcp*[h]{while experiment runs}){$t \gets 0$ \KwTo $t_\mathrm{max}$}{
            $ \mathbf{o}_t \gets \Observe()$ \;
            \BlankLine
            \ForEach{ $\mathrm{act} \in \{ \mathrm{pick}, \mathrm{place}\}$} 
            {
            $\mathbf{T} = \{ (u_1, v_1), \dots , (u_{k}, v_{k})\} \gets \TopAnalysis(\mathcal{Q}_\mathrm{act}((u,v)|\mathbf{o}_t))$ \tcp*[r]{$k$ local maxima}
			
            
			$\hat{p}_{\mathrm{act}} \gets \AmbiguityMeasure (\mathbf{T})$ \;
			\BlankLine
			\uIf(\tcp*[h]{if ambiguous}){ $\hat{p}_{\mathrm{act}} \leq p_{\mathrm{act}}^\mathrm{thr}$} {
			        $\mathbf{a}_t \gets \QueryTeacher(\mathbf{T})$\;
					$  \Act(\mathbf{a}_t), \Dataset \gets \Dataset \cup (\mathbf{o}_t, \mathbf{a}_t)$\tcp*[r]{adding user input to the Dataset}
         
			}
			\Else(\tcp*[h]{if not ambiguous}){
			$ \Act(\argmax_{(u, v) \in \mathbf{T}}  \mathcal{Q}_\mathrm{act}((u,v)|\mathbf{o}_t))$ \;
            
            \If (\tcp*[h]{if teacher corrects}) {$\mathbf{a}_\mathrm{corr} \gets \ObserveCorrection() \neq \varnothing$}{
            $\Dataset \gets \Dataset \cup (\mathbf{o}_t, \mathbf{a}_\mathrm{corr})$\;
            }
			}
            $p_{\mathrm{act}}^\mathrm{thr} \gets \Threshold (p_{\mathrm{act}}^\mathrm{thr}), \mathcal{Q}_\mathrm{act} \gets \Train (\Dataset)$ \tcp*[r]{update the model with new data}
			}
			}
	}
    \captionsetup{belowskip=0pt}
    \setlength{\belowcaptionskip}{-8pt}
    \addtolength{\textfloatsep}{-0.2in}
    \caption{PARTNR} 
    \label{alg:partnr}
\end{algorithm}

We evaluated the performance of the proposed method in a simulated table-top pick and place task, which is shown in \Cref{fig:env}.
This task is very similar to tasks from \cite{shridhar2021cliport, zeng2021transporter, zeng2022socraticmodels}.
The goal of this task is to execute language commands in the form ``\textit{Pick the [pick color] box and place it in the [place color] bowl.}'', where the pick and place colors are sampled at the beginning of an episode, based on the colors of the objects present in the scene.
The task is simulated using the PyBullet simulator \cite{coumans2021} and the implementation is adapted from \cite{zeng2022socraticmodels}.
At the beginning of each episode, three boxes and three bowls are placed at random locations on the table. 
Similar to \cite{shridhar2021cliport}, the colors of the objects are either sampled from the color set of $\mathcal{C}_\text{all} \cup \mathcal{C}_\text{seen}$ or the color set $\mathcal{C}_\text{all} \cup \mathcal{C}_\text{unseen}$, where $\mathcal{C}_\text{all} =$ \{red, blue, green\}, $\mathcal{C}_\text{seen} =$ \{yellow, brown, gray, cyan\}, and $\mathcal{C}_\text{unseen} =$ \{orange, purple, pink, white\}.
The set of seen colors is used for offline training, while both sets are used for evaluation and interactive learning, to simulate a domain shift.

As a baseline, the CLIPort variant used in \cite{zeng2022socraticmodels, ahn2022can} is employed and trained on a dataset consisting of demonstrations from a scripted expert.
We also trained CLIPort models using the PARTNR algorithm with the same architecture as the baseline interactively, as described in \Cref{alg:partnr}.
The interactive models are initially trained offline and updated interactively while executing the task.
To have a fair comparison between the baseline and the interactive models, both have the same number of total demonstrations and a total number of model updates.
Since real-life demonstrations are never perfect, we also evaluated the method with noisy demonstrations.

We follow \cite{shridhar2021cliport} and evaluate each model in 100 episodes consisting of three pick-and-place commands.
The percentage of successfully performed pick and place commands is used as evaluation metric.
The results in \Cref{tab:results} show that the PARTNR algorithm improves the baseline performance, both in the in-distribution and out-of-distribution scenarios (seen and unseen case).
The improvement in the unseen case indicates that by collecting on-policy data, the methods improves robustness against domain shifts.
The on-policy data also enables to learn to recover from failure states, which is not possible in the offline baseline.
Interestingly, both the baseline and PARTNR performance improved substantially when adding noise to the pick and place demonstrations from the scripted expert.
To be noted, the final performance is lower than obtained with the original CLIPort model in \cite{shridhar2021cliport}.
Most likely, this is due to the usage of a simplified variant, a lower number of data augmentations and a lower number of camera perspectives.
However, this is not relevant as the main focus here is to make a comparison against a non-interactive baseline, and not to obtain optimal performance.

\begin{minipage}{\textwidth}
    \begin{minipage}[c]{0.19\textwidth}
        \centering
        \includegraphics[width=\linewidth]{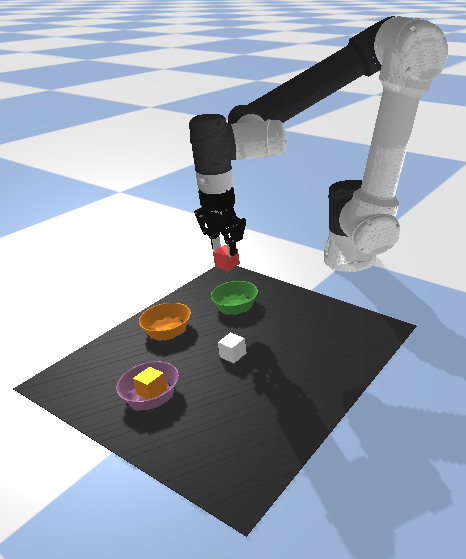}
        \captionof{figure}{\scriptsize{The \textit{put-blocks-in-bowls} task.}}
        \label{fig:env}
    \end{minipage}
    \hfill
    \begin{minipage}[c]{0.79\textwidth}
        \centering
        \setlength\tabcolsep{2.3pt}
        \scriptsize
        \begin{tabular}{lccccccccc}
            \toprule
            & & \multicolumn{4}{c}{\begin{tabular}[c]{@{}c@{}}put-blocks-in-bowls \\ seen-colors\end{tabular}}   & \multicolumn{3}{c}{\begin{tabular}[c]{@{}c@{}}put-blocks-in-bowls \\ unseen-colors\end{tabular}}    \\
            \cmidrule(lr){3-6} \cmidrule(lr){7-9}\\[-5pt]
           algorithm & data split & 500 & 1000 & 1000 noisy & 1500 & 500 & 1000 & 1500  \\ 
            \midrule
            Baseline & 100\% off & 28.3 & 51.7 & 82.7 & 62.7 & 19.0 & 22.0 & 16.7 \\
            \rowcolor[rgb]{0.792,1,0.792} PARTNR & 50\% off + 50\% int & \textbf{30.3} & \textbf{57.3} & \textbf{91.0} & \textbf{80.3} & \textbf{ 30.7} & \textbf{53.0} & \textbf{78.3} \\
            PARTNR (80\% data) & 50\% off + 30\% int & 28.0 & 39.3 & 77.7 & 68.0 & 20.3 & 28.3 & 57.3 \\
            \bottomrule
        \end{tabular}
        \captionof{table}{\scriptsize{
            The performance of the PARTNR algorithm is evaluated against the performance of the non-interactive baseline (CLIPort variant).
            Here the success rate (\%) is shown for a number of demonstrations, i.e., 500, 1000 and 1500.
            The data split indicates the percentage of demonstrations that were obtained offline (off) and interactively (int), so in the last row, 20\% fewer demonstrations were collected.
            Since real demonstrations are often noisy, we also evaluated both methods with noise ($\sim \mathcal{N}(0, 3^2)$ pixels) added to the pick and place locations (1000 noisy).}
        }
        \label{tab:results}
    \end{minipage}
\end{minipage}

\section{Conclusions and Outlook}

This work introduced PARTNR, an interactive learning algorithm for resolving ambiguities in pick-and-place tasks.
The PARTNR algorithm improves the baseline performance, both in the in-distribution and out-of-distribution scenarios.
Furthermore, sampling efficiency is improved (even up to 20\% more data-efficient), since demonstrations are only collected when needed, based on the user-specified sensitivity.
In the future, we plan to evaluate PARTNR with the original CLIPort baseline as well and to further address the epistemic uncertainty of the model, e.g., through an ensemble approach.
Also, we wish to extend the method with sequence prediction and feedback control.
Finally, we plan to monitor the human cognitive load in a real-world participant study.

\section*{Acknowledgments}

This work was supported by the European Union’s H2020 project Open Deep Learning Toolkit for Robotics (OpenDR) under grant agreement \#871449 and by the ERC Stg TERI, project reference \#804907.

\medskip
\small

\bibliographystyle{plain}
\bibliography{bib.bib}

\appendix

\section{Recovering from mistake}

\begin{figure}[!htb]
    \centering
    \begin{subfigure}[b]{0.24\textwidth}
        \centering
        \includegraphics[width=\textwidth]{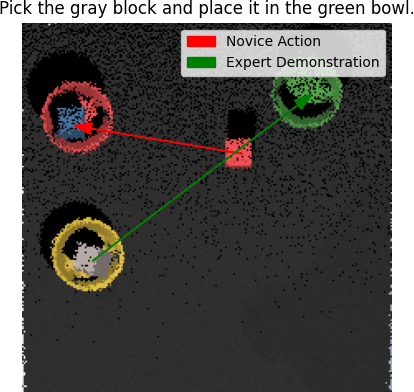}
        \caption{}
    \end{subfigure}
    \hfill
    \begin{subfigure}[b]{0.24\textwidth}
        \centering
        \includegraphics[width=\textwidth]{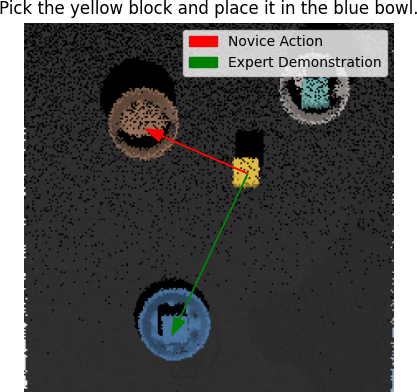}
        \caption{}
    \end{subfigure}
        \hfill
    \begin{subfigure}[b]{0.24\textwidth}
        \centering
        \includegraphics[width=\textwidth]{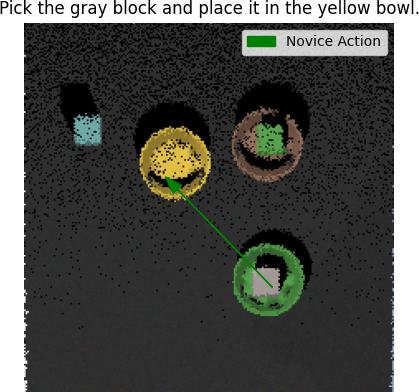}
        \caption{}
    \end{subfigure}
    \hfill
    \begin{subfigure}[b]{0.24\textwidth}
        \centering
        \includegraphics[width=\textwidth]{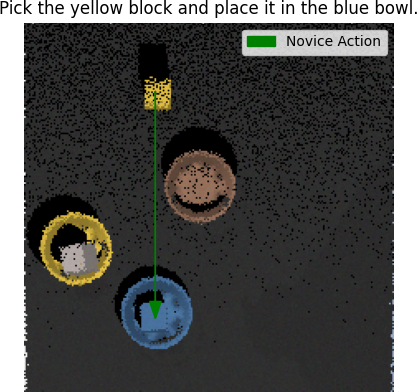}
        \caption{}
    \end{subfigure}
    \caption{Example of learning how to recover thanks to on-policy data collection.
    The figures \textbf{(a)} and \textbf{(b)} show demonstrations for failure states, i.e., in \textbf{(a)} the block to be picked is already in another bowl, and in \textbf{(b)} there is already a block in the blue bowl.
    Such demonstrations of failure states are collected only in the interactive case.
    Figures \textbf{(c)} and \textbf{(d)} show that the novice can learn to recover from such failure states. }
    \label{fig:recov}
\end{figure}

Because the PARTNR algorithm collects data from the state distribution induced by the novice policy, it can learn to recover after mistakes, while this is not the case when learning offline from expert demonstrations.
That is to say, the expert does not make any mistakes and therefore failure states are not visited by the expert policy.
An example of such a recovery learned interactively is shown in \Cref{fig:recov}.

\section{PARTNR framework: Detailed algorithm}

A detailed version of the PARTNR algorithm is shown in \Cref{alg:partnr_detailed}.
For computing the sensitivity or optionally specificity \cite{chu1999introduction}, we also need to keep track of the true positives ($\mathrm{TP}$), true negatives ($\mathrm{TN}$), false positives ($\mathrm{FP}$) and false negatives ($\mathrm{FN}$).
The definition of positives and negatives is shown in \Cref{tab:posneg}.

\begin{table}[!htb]
    \centering
    \caption{Definition of positives and negatives. In the ideal case, the teacher is only queried in the case that the robot's action would result in a failure (true positive).
    }
    \label{tab:posneg}
    \begin{tabular}{@{}cccc@{}}
        & & \multicolumn{2}{c}{Human input was necessary} \\
        \multirow{4}{*}{\ctab{\rotatebox[origin=c]{90}{Ambiguous}}}
        & & \multicolumn{1}{c}{True} & \multicolumn{1}{c}{False}\\
        \cmidrule(l){3-4} 
        & \multicolumn{1}{l|}{True} & True Positive (TP) & False Positive (FP) \\
        & \multicolumn{1}{l|}{False} & False Negative (FN) & True Negative (TN) \\
    \end{tabular}
\end{table}

\begin{algorithm}[t]
\DontPrintSemicolon
	\SetKwFunction{Observe}{Observe}
	\SetKwFunction{CLIPort}{\cliport}
	\SetKwFunction{Train}{Train}
	\SetKwFunction{QueryTeacher}{QueryTeacher}
	\SetKwFunction{TopAnalysis}{TopAnalysis}
	\SetKwFunction{AmbiguityMeasure}{AmbiguityMeasure}
	\SetKwFunction{Threshold}{UpdateThreshold}
	\SetKwFunction{ObserveCorrection}{ObserveCorrection}
	\SetKwFunction{UpdateFlags}{UpdateFlags}
	\SetKwFunction{Act}{Act}
	\SetKwInOut{Input}{input}\SetKwInOut{Output}{output}
 	\Input{$\Dataset^\mathrm{init}$ \tcp*[r]{initial demonstrations}}
	\Output{$\mathcal{Q}_\mathrm{pick}, \mathcal{Q}_\mathrm{place}$ \tcp*[r]{pick and place value functions} } 
	\BlankLine
	\Begin{
	
		$\Dataset \gets \Dataset^\mathrm{init}$ \tcp*[r]{initial Dataset}
		$\mathcal{Q}_\mathrm{pick}, \mathcal{Q}_\mathrm{place} \gets \Train (\Dataset^\mathrm{init}) $\;
        $ \mathrm{TP}, \mathrm{TN}, \mathrm{FP}, \mathrm{FN} \gets \varnothing$\;
        $p_{\mathrm{pick}}^\mathrm{thr} , p_{\mathrm{place}}^\mathrm{thr}            \gets \texttt{init}()  $\;
		\BlankLine
		\For(\tcp*[h]{while experiment runs}){$t \gets 0$ \KwTo $t_\mathrm{max}$}{
            $ \mathbf{o}_t \gets \Observe()$ \;
            \BlankLine
            \ForEach{ $\mathrm{act} \in \{ \mathrm{pick}, \mathrm{place}\}$} 
            {
            $\mathrm{isUpdated} \gets \texttt{false}$\;
            $\mathbf{T} = \{ (u_1, v_1), \dots , (u_{k}, v_{k})\} \gets \TopAnalysis(\mathcal{Q}_\mathrm{act}((u,v)|\mathbf{o}_t))$ 
			
			$\mathbf{a}_\mathrm{max} \gets  \argmax_{(u, v) \in \mathbf{T}}  \mathcal{Q}_\mathrm{act}(u, v)$\;
            
			$\hat{p}_{\mathrm{act}} \gets \AmbiguityMeasure (\mathbf{T})$ \;
			\BlankLine
			\uIf(\tcp*[h]{if ambiguous}){ $\hat{p}_{\mathrm{act}} \leq p_{\mathrm{act}}^\mathrm{thr}$} {
			        $\mathbf{a}_t \gets \QueryTeacher(\mathbf{T})$\;
					$\Dataset \gets \Dataset \cup (\mathbf{o}_t, \mathbf{a}_t)$\tcp*[r]{adding user input to the Dataset}
                    $\mathrm{isUpdated} \gets \texttt{true}$\;
					\If {$\mathbf{a}_t ==\mathbf{a}_\mathrm{max}$ }{
					$ \mathrm{FP} \gets \mathrm{FP} \cup {t}$ \tcp*[r]{adding False Positive flag}
					\Else{
					$ \mathrm{TP} \gets \mathrm{TP} \cup {t}$ \tcp*[r]{adding True Positive flag}
            }
            }
            $ \Act(\mathbf{a}_t)$ \;
			}
			\Else(\tcp*[h]{if not ambiguous}){
			$ \Act(\mathbf{a}_\mathrm{max})$ \;
            
            \If (\tcp*[h]{if teacher corrects}) {$\mathbf{a}_\mathrm{corr} \gets \ObserveCorrection() \neq \varnothing$}{
            $\Dataset \gets \Dataset \cup (\mathbf{o}_t, \mathbf{a}_\mathrm{corr})$\;
            $\mathrm{isUpdated} \gets \texttt{true}$\;
            $ \mathrm{FN} \gets \mathrm{FN} \cup {t}$ \tcp*[r]{adding False Negative flag}
            }
            \Else{
            $ \mathrm{TN} \gets \mathrm{TN} \cup {t}$ \tcp*[r]{adding True Negative flag}
            }
			}
            $p_{\mathrm{act}}^\mathrm{thr} \gets \Threshold (p_{\mathrm{act}}^\mathrm{thr}, \mathrm{TP}, \mathrm{TN}, \mathrm{FP}, \mathrm{FN})$
			\BlankLine
			\If {$\mathrm{isUpdated}$} {

            $\mathcal{Q}_\mathrm{act} \gets \Train (\Dataset)$ \tcp*[r]{update the model with new data}
            }
            
			}

			}

	}
    \caption{PARTNR - detailed algorithm}
    \label{alg:partnr_detailed}
\end{algorithm}

\section{Adaptive Threshold: More details}\label{sec:threshold}

A detailed version of the adaptive threshold algorithm is shown in \Cref{alg:at}.
Here, the number of true positives, true negatives, false positives, and false negatives are counted over a window ($k_\mathrm{TP}, k_\mathrm{TN}, k_\mathrm{FP}, k_\mathrm{FN}$, respectively).
Subsequently, the sensitivity can be estimated ($\hat{s}$) \cite{chu1999introduction}.
Finally, the threshold $p^\mathrm{thr}$ is updated proportionally to the error between the desired and estimated sensitivity.
In our experiments, we used the following values: $p_0^\mathrm{thr} = 0.5$, $s_\mathrm{des} = 0.9$, $w_n = 50$ and $a = 0.005$.

\begin{algorithm}[!htb]
	\DontPrintSemicolon
	\fontsize{10pt}{10pt}\selectfont
	\SetKwInOut{Input}{input}
	\SetKwInOut{Output}{output}
	\SetKwFunction{Threshold}{UpdateThreshold}
	\SetKwFunction{Filter}{MovHorCnt}
	\Input{Initial threshold $p_0^\mathrm{thr} $, Desired sensitivity $s_{\mathrm{des}}$, Window length $w_n$, Adaptation rate $a$.}
	\Output{threshold $p^\mathrm{thr}$}
	
	\Begin{
	$k_\mathrm{TP}, k_\mathrm{TN}, k_\mathrm{FP}, k_\mathrm{FN} \gets \Filter ( w_n, \mathrm{TP}, \mathrm{TN}, \mathrm{FP}, \mathrm{FN}) $\tcp*[r]{Counting occurrence in the window $w_n$} 
	$\hat{s} \gets \frac{k_\mathrm{TP}}{k_\mathrm{TP} + k_\mathrm{FN}}$\;
	$p^\mathrm{thr} \gets p_0^\mathrm{thr} - a \cdot ( s_\mathrm{des}-\hat{s} )$\; 
    }
    \caption{\protect\Threshold }
    \label{alg:at}
    
\end{algorithm}

\section{Ambiguity Measure: Example and visualization}\label{sec:ambiguity}

\begin{figure}
    \centering
    \begin{subfigure}[b]{0.32\textwidth}
        \centering
        \includegraphics[width=0.9\textwidth]{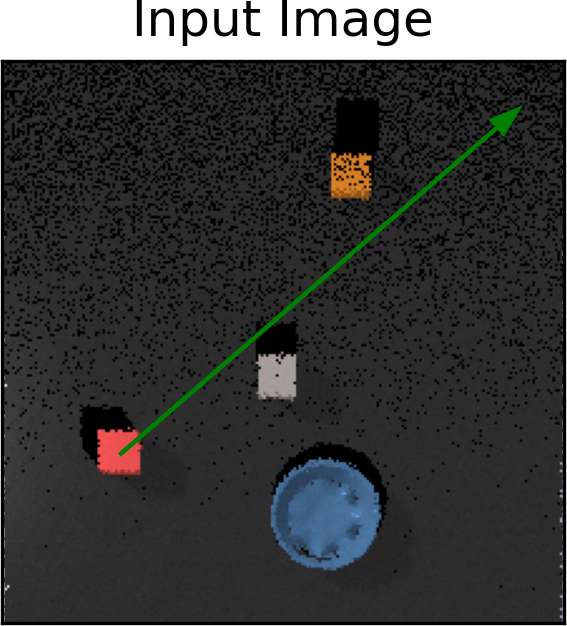}
        \caption{}
    \end{subfigure}
    \hfill
    \begin{subfigure}[b]{0.32\textwidth}
        \centering
        \includegraphics[width=0.9\textwidth]{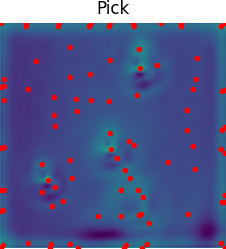}
        \caption{}
    \end{subfigure}
    \hfill
    \begin{subfigure}[b]{0.32\textwidth}
        \centering
        \includegraphics[width=0.9\textwidth]{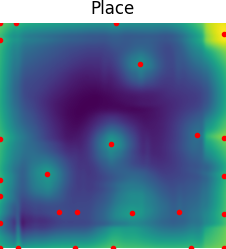}
        \caption{}
    \end{subfigure}
    \\
    \begin{subfigure}[b]{0.49\textwidth}
        \centering
        \includegraphics[width=0.9\textwidth]{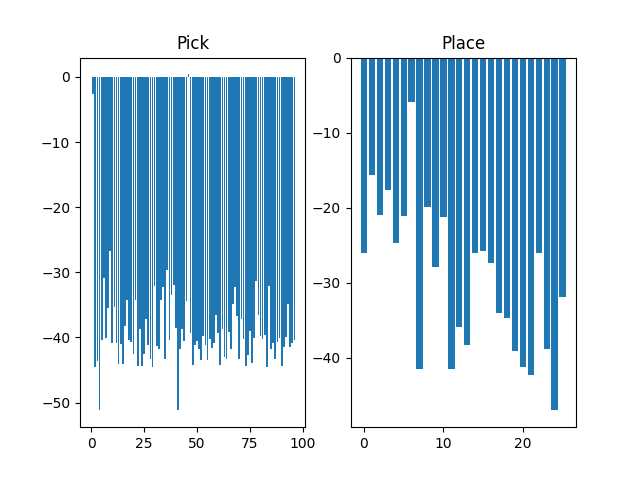}
        \caption{}
    \end{subfigure}
    \hfill
    \begin{subfigure}[b]{0.49\textwidth}
        \centering
        \includegraphics[width=0.9\textwidth]{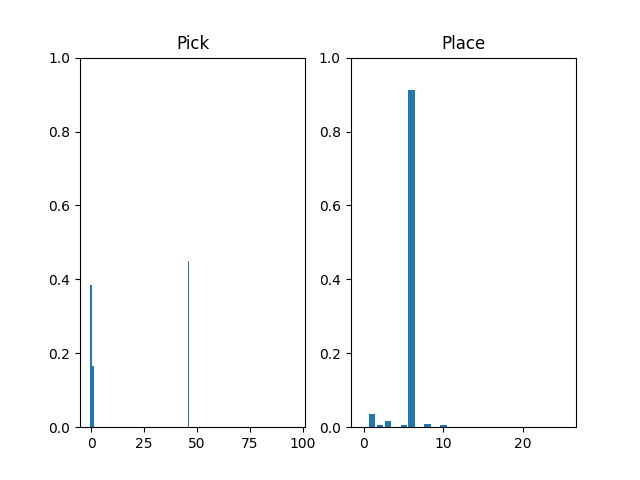}
        \caption{}
    \end{subfigure}
    \\
    \begin{subfigure}[b]{0.49\textwidth}
        \centering
        \includegraphics[width=0.66\textwidth]{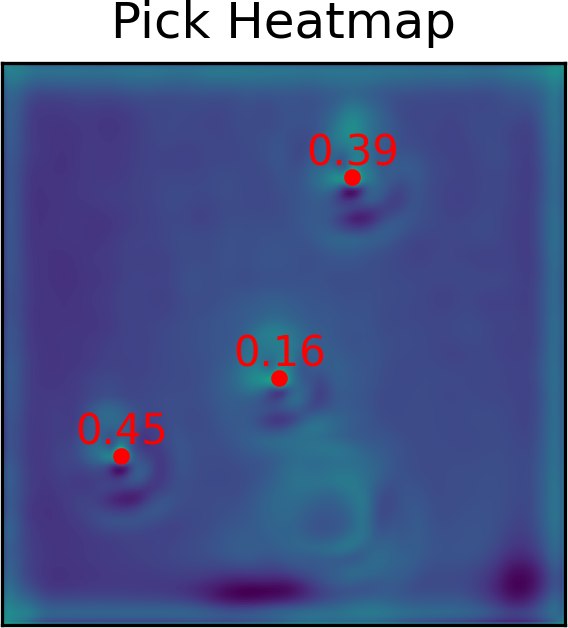}
        \caption{}
    \end{subfigure}
    \hfill
    \begin{subfigure}[b]{0.49\textwidth}
        \centering
        \includegraphics[width=0.66\textwidth]{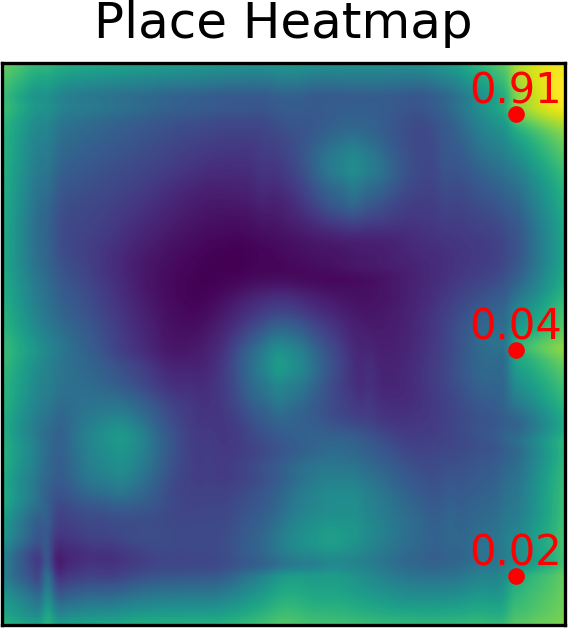}
        \caption{}
    \end{subfigure}
    \caption{This figure shows a visual example of obtaining the ambiguity measure. The input image together with the correct action (green arrow) is shown in \textbf{(a)}. Here, the language command is: `\textit{Pick the red block and place it on the top right corner}''. With $\texttt{TopAnalysis}$, we obtain local maxima $\mathbf{T}$, which are shown in \textbf{(b)} and \textbf{(c)} for the pick and place poses, respectively. The corresponding values are shown in \textbf{(d)}. After normalization using the $\mathrm{softmax}$ function, we obtain \textbf{(e)}. The local maxima with a normalized value greater than 0.01 are shown in \textbf{(f)} and \textbf{(g)} for the pick and place poses, respectively. The maximum of the normalized values is used as ambiguity measure.}
    \label{fig:ambiguity_measure}
\end{figure}

Figure \Cref{fig:ambiguity_measure} shows a visual example of how the ambiguity measure is obtained.

\end{document}